\documentclass[conference]{IEEEtran}
\usepackage[T1]{fontenc}
\usepackage{cite}
\usepackage{amsmath,amssymb,amsfonts}
\usepackage{graphicx}
\usepackage[caption=false,font=footnotesize]{subfig}
\usepackage{booktabs}
\usepackage{tabularx}
\usepackage[table]{xcolor}
\usepackage[hidelinks]{hyperref}

\definecolor{methodgreen}{HTML}{E7F1E5}
\definecolor{drivingblue}{HTML}{DBEAF4}
\definecolor{burdenorange}{HTML}{F9E6D4}
\definecolor{lightmauve}{rgb}{0.86, 0.82, 1.0}
\begin{document}

\title{Value and Policy Calibration from Execution Feedback for Socially Compatible Driving}

\author{
\IEEEauthorblockN{Sen Shen$^{1}$, Ning Liu$^{2}$, Xiangrui Kong$^{2}$, Yingwei Li$^{1}$, Dongkun Han$^{1*}$}
\IEEEauthorblockA{$^{1}$ Department of Mechanical and Automation Engineering, The Chinese University of Hong Kong, Hong Kong, China.}
\IEEEauthorblockA{$^{2}$ Department of Electrical, Electronic and Computer Engineering, The University of Western Australia,
Perth, Australia.}
\IEEEauthorblockA{senshen@link.cuhk.edu.hk, \{ning.liu, xiangrui.kong\}@research.uwa.edu.au, \{ywli, dkhan\}@mae.cuhk.edu.hk}
}

\maketitle

\begin{abstract}
Autonomous vehicle (AV) decisions at intersections affect the delay, braking, and motion smoothness of surrounding human-driven vehicles (HDVs). Learning socially compatible behavior requires connecting imagined interaction outcomes with feedback from executed actions. We propose a world-model reinforcement learning framework combines imagined policy improvement with direct value and policy calibration from finalized interactions, balancing AV progress with HDV interaction burden. An interaction-aware latent model supports imagination, while off-policy correction and replay updates enable observed rewards and transitions to supervise both the critic and actor. Simulation experiments at an unsignalized intersection achieve a 100\% success and reductions in following-HDV acceleration variation of 65.5\% and 44.0\% relative to PIWM and DreamerV3, respectively. Ablations identify complementary roles for value calibration in reducing HDV burden and policy calibration in improving passage efficiency. Tests under out-of-distribution traffic shifts further show lower ego jerk and HDV acceleration variation than PIWM at comparable passage times.
\end{abstract}

\section{Introduction}
\label{sec:introduction}

\begin{figure}[!t]
    \centering
    \subfloat[]{%
        \includegraphics[width=0.49\columnwidth]{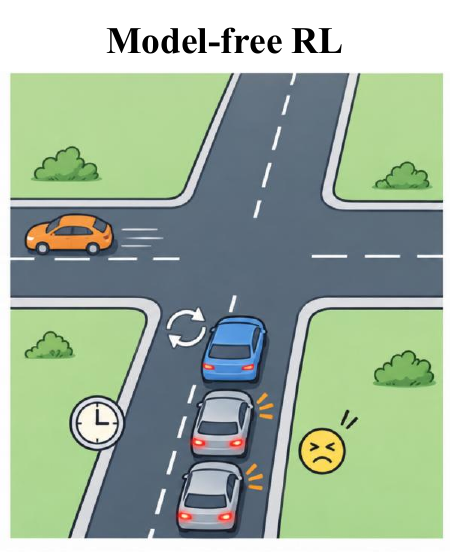}%
        \label{fig:interaction_a}%
    }\hfill
    \subfloat[]{%
        \includegraphics[width=0.49\columnwidth]{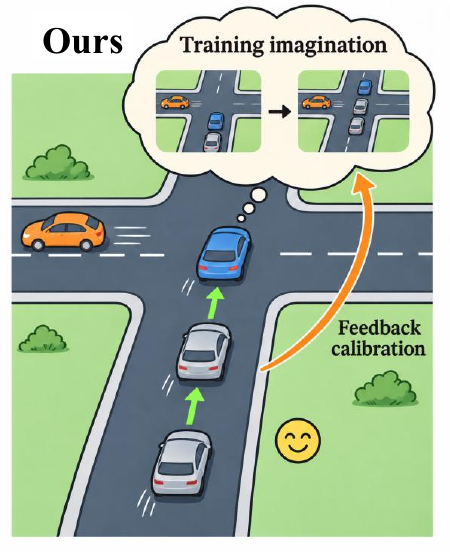}%
        \label{fig:interaction_b}%
    }
    \caption{Conceptual comparison: (a) waiting behind a hesitant AV;
    (b) smoother following through training-time imagination and
    executed-feedback calibration.}
    \label{fig:interaction_teaser}
\end{figure}

At mixed-traffic intersections, the decisions of an autonomous vehicle (AV) affect both its own progress and the behavior of surrounding human-driven vehicles (HDVs)~\cite{sadigh2016planning}. The timing of entry into a conflict area can affect when other vehicles brake or wait, while prolonged hesitation may cause following vehicles to stop and restart repeatedly (Fig.~\ref{fig:interaction_teaser}\subref{fig:interaction_a}). Evaluating AV driving quality therefore requires considering both the AV's performance and the interaction costs borne by surrounding road users~\cite{sun2018courteous}.

The interests of surrounding vehicles do not always align. Yielding to traffic from a conflicting direction may increase the waiting time of following vehicles, while sustained low speeds may introduce additional speed fluctuations. Courtesy therefore involves more than simply driving slowly or stopping more often~\cite{wang2021socially}; it requires balancing the AV's objectives with those of other road users~\cite{schwarting2019social}. We characterize the interaction burden on HDVs using delay, braking burden, and acceleration variation, and study how a single AV can balance its own efficiency with these burdens while maintaining safe passage.

Multi-agent reinforcement learning and hybrid planning methods have been used to coordinate multiple agents as they interact with dynamic environments~\cite{liu2025cooperative,liu2026vorl}. In mixed traffic, prior reinforcement learning research has used centralized intersection control to direct vehicles to yield the right of way, improving traffic efficiency and fairness~\cite{han2024multi}. World-model reinforcement learning learns environment dynamics and predicts future states and
rewards, enabling driving policies to be optimized using imagined trajectories~\cite{li2024think2drive}. These advances provide a basis for combining socially compatible interaction objectives with policy learning through imagined interactions.

However, specifying an objective that accounts for HDV burden leaves open how interaction outcomes should inform policy learning. In learning from imagined trajectories, value targets depend on both model predictions and value estimates~\cite{buckman2018sample}. Executed interactions provide observed delay and braking feedback, but replay trajectories are collected under behavior policies from different training stages. Reusing them to update the current policy therefore requires accounting for the resulting policy mismatch~\cite{espeholt2018impala}. We investigate how finalized feedback from completed interactions can directly support value and policy updates alongside continued learning from imagined trajectories.

To this end, we propose a world-model reinforcement learning framework that calibrates value and policy learning using feedback from executed interactions (Fig.~\ref{fig:interaction_teaser}\subref{fig:interaction_b}). The framework uses individual vehicle representations and interaction prediction to generate imagined trajectories for policy learning. Alongside these imagined rollouts, we use completed simulation trajectories whose interaction feedback has been finalized. Rewards from the executed trajectories are used to construct value targets and policy advantages, allowing observed outcomes to directly guide both actor and critic updates. We apply off-policy correction to account for differences between historical behavior policies and the current policy when reusing these trajectories. Coordinated replay and updates combine learning from executed and imagined experience to balance AV progress with the interaction burden on surrounding HDVs.

The main contributions are as follows:
\begin{itemize}
    \item A world-model reinforcement learning framework balances AV progress with HDV delay, braking burden, and acceleration variation at mixed-traffic intersections.
    \item By combining finalized execution feedback with off-policy correction and replay updates, the training method jointly calibrates value and policy learning.
\end{itemize}

\section{Related Work}
\label{sec:related_work}

\subsection{Socially Compatible Driving.}
Cooperative driving methods account for differences in how human drivers respond to AV actions. Belief-based reinforcement learning estimates other drivers' cooperation levels when learning to merge in dense traffic~\cite{bouton2019cooperation}. SymCoDrive learns cooperative AV behavior while accounting for human drivers' utility~\cite{toghi2021cooperative}, and GamePlan uses observable driving styles to assign priorities at intersections, roundabouts, and merges~\cite{chandra2022gameplan}. These studies connect human behavior modeling and social preferences with interaction decisions. Our work considers a single AV and focuses on using observed interaction outcomes to balance its progress with the burden on crossing and following HDVs.

\subsection{World Models and Policy Learning.}
World-model and off-policy reinforcement learning provide complementary ways to use executed experience in policy optimization. DreamerV3 learns through latent imagination and also trains its critic on replayed environment
experience~\cite{hafner2025mastering}. MBPO uses short model rollouts starting from replayed environment states~\cite{janner2019trust}, while ACC uses recent on-policy returns to adapt critic bias correction~\cite{dorka2022adaptively}. Shared experience replay combines replayed and on-policy data while accounting for differences between behavior and learning policies~\cite{schmitt2020off}. Training adaptation has also been studied through dynamic update-to-data ratios guided by held-out world-model performance and selective model or policy updates during adaptation to new tasks~\cite{wang2025adawm}. Our framework uses finalized feedback from completed simulation interactions to calibrate both value and policy learning, coordinating replay and updates to account for AV progress and HDV interaction burden.

\section{Problem Formulation}
\label{sec:problem}

We consider speed control of a single autonomous vehicle (AV) at a mixed-traffic intersection, balancing its progress with the interaction burden on surrounding human-driven vehicles (HDVs). Let $h_t$ denote the available motion history of the ego and surrounding vehicles at time $t$, including positions, velocity components, headings, and road information. Given $h_t$, a policy $\pi(a_t\mid h_t)$ selects a target speed $a_t$ from a discrete set $\mathcal A$. The realized motion follows the simulator's vehicle dynamics, car-following rules, and traffic constraints.

Over all HDVs, delay $D$ sums SUMO travel-time loss and waiting from scheduled departure to network entry. With $\Delta t=0.1\,\mathrm{s}$ and acceleration $a$ in $\mathrm{m/s^2}$, excess braking $B$ sums $\Delta t\max(0,-a-2)$ over valid samples. Acceleration variation $J$ sums $|a_t-a_{t-1}|$ over consecutive valid samples of each HDV. These costs enter the reward as
\begin{equation}
\begin{aligned}
r_{t+1}={}&r^{\mathrm{ego}}_{t+1}
-w_D\Delta D_{t+1}\\
&-w_B\Delta B_{t+1}
-w_J\Delta J_{t+1}.
\end{aligned}
\label{eq:method_reward}
\end{equation}
Here, $r^{\mathrm{ego}}_{t+1}$ includes progress, successful traversal, elapsed time, and safety events. The increments $\Delta D_{t+1}$, $\Delta B_{t+1}$, and $\Delta J_{t+1}$ measure the burden added during the current transition, and $w_D,w_B,w_J$ are fixed training weights. The reward evaluates both ego performance and the burden on surrounding vehicles when choosing whether to wait or proceed.

Since actions also affect subsequent traffic states, we maximize the expected discounted return over the AV control period:
\begin{equation}
\max_\pi \mathbb E_\pi \sum_{t=0}^{T-1}\gamma^t \bar r_{t+1}.
\label{eq:problem_objective}
\end{equation}

Here, $\gamma$ is the discount factor, $\bar r_{t+1}$ the finalized reward, and $T$ the AV control horizon. For every method, evaluation tracks all scheduled HDVs until all vehicles leave or simulation time reaches $60\,\mathrm{s}$, whichever comes first. Discounted post-control HDV costs enter the terminal reward $\bar r_T$ (Section~\ref{sec:method_value}).

\begin{figure*}
    \centering
    \includegraphics[width=1.0\linewidth]{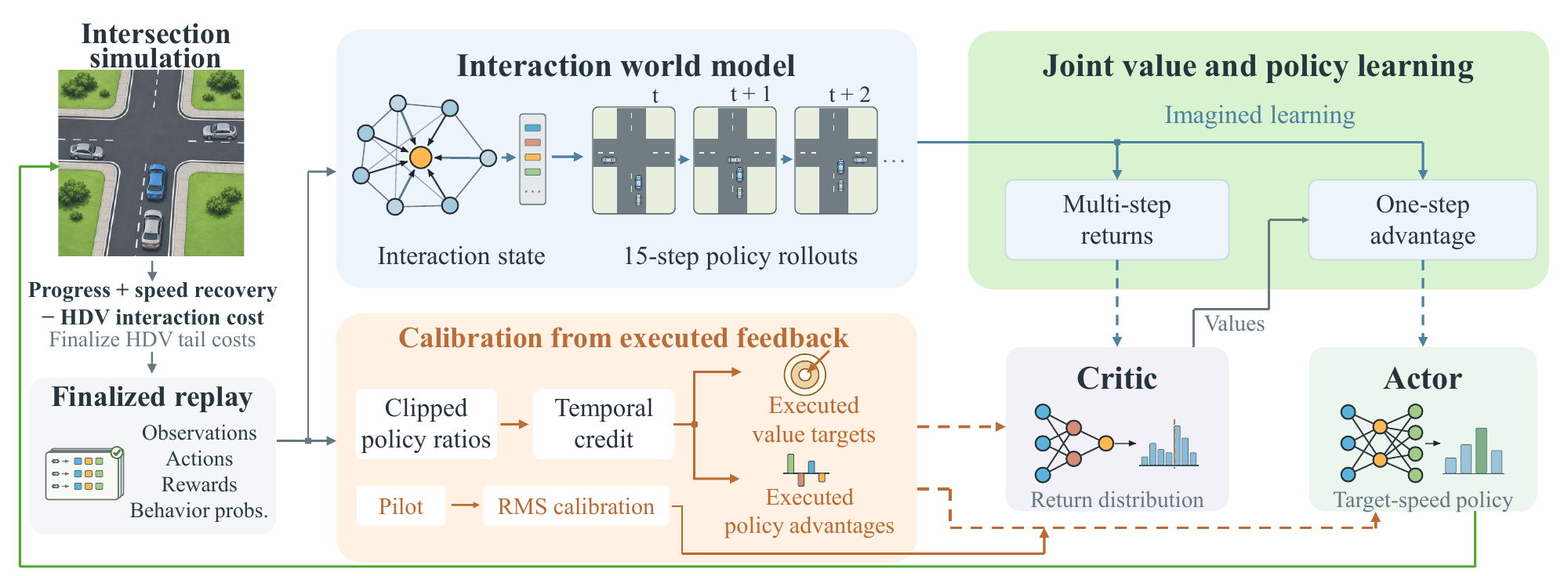}
    \caption{Overview of the proposed framework. Executed simulation experience trains the world model and calibrates the actor and critic. Both networks also learn from imagined rollouts, with the calibrated critic supporting action evaluation.}
    \label{fig:framework}
\end{figure*}

\section{Method}
\label{sec:method}

\subsection{Overview}
\label{sec:method_overview}

Fig.~\ref{fig:framework} combines world-model imagination with executed-feedback calibration. Individual vehicle representations and interaction attention encode history $h_t$ into latent state $x_t$ to predict motion, rewards, and continuation. The actor $\pi_\theta(a_t\mid x_t)$ selects target speeds, and the critic $V_\psi(x_t)$ estimates returns. Finalized execution feedback calibrates both networks, grounding imagined action evaluation in observed outcomes and directly updating executed action probabilities. Imagined rollouts provide multistep critic targets and one-step actor advantages evaluated by the calibrated critic.

\subsection{Value Calibration from Executed Interactions}
\label{sec:method_value}

We record observations, actions, behavior probabilities, and interaction costs. The final stage uses $(w_D,w_B,w_J)=(0.2,0.2,0.05)$. On the outbound lane, the speed-recovery penalty applies if no leader is detected within $100\,\mathrm{m}$ or its reported gap satisfies $g\geq 2.5+v+v^2/6$, with $g$ in metres and AV speed $v$ in m/s. The penalty is $0.1\max(v_{\rm lim}-v,0)/v_{\rm lim}$, where $v_{\rm lim}=\min(9\,\mathrm{m/s},v_{\rm lane})$. Before replay insertion, we reconcile delay with vehicle-exit records and fold discounted post-control HDV costs into the terminal reward. Replay retains the $T$ control steps and finalized rewards $\bar r_{t+1}$. Motion histories are re-encoded into consecutive states $x_t,x_{t+1}$ for training.

To regulate historical actions from different behavior policies, we use the clipped probability ratio
\begin{equation}
\rho_t=\min\left(
1,\frac{\pi_\theta(a_t\mid x_t)}{\mu_t(a_t)}
\right),
\label{eq:method_ratio}
\end{equation}
where $\mu_t(a_t)$ is the recorded collection probability. The numerator uses the current policy and re-encoded state, making $\rho_t$ a bounded calibration weight: actions less likely under the policy are downweighted, and other weights are capped at one.

Executed feedback corrects the current value estimate through the temporal-difference (TD) error
\begin{equation}
\begin{aligned}
\delta_t={}&\bar r_{t+1}
+\gamma(1-d_{t+1})V_\psi(x_{t+1})-V_\psi(x_t),
\end{aligned}
\label{eq:method_td}
\end{equation}
where $d_{t+1}=1$ marks the end of AV control and disables bootstrapping. Finalized HDV burden thus enters the value target through $\bar r_{t+1}$, including at termination.

The errors propagate backward with a clipped trace~\cite{espeholt2018impala} over valid segments with known behavior probabilities:
\begin{equation}
e_t=\rho_t\delta_t
+\gamma(1-d_{t+1})\lambda\rho_t e_{t+1},
\label{eq:method_value_correction}
\end{equation}
where $\lambda=0.95$ controls temporal decay. The executed value target is
\begin{equation}
y_t=V_\psi(x_t)+e_t.
\label{eq:method_value_target}
\end{equation}
The subsequent correction is zero at segment boundaries. Valid transitions without behavior probabilities retain a one-step target from the finalized reward and next-state value.

The critic combines executed targets $y_t$ with multistep imagined returns:
\begin{equation}
\mathcal L_V=
\mathcal L_V^{\mathrm{imag}}
+\beta\mathcal L_V^{\mathrm{exec}}.
\label{eq:method_value_loss}
\end{equation}
Here, $\beta$ weights executed supervision. Both terms fit a return distribution by likelihood loss, with regularization toward a slowly updated value network. Executed targets and latent inputs are detached, so calibration updates the critic without backpropagating into the world model.

\begin{table*}[!t]
    \centering
    \caption{Performance comparison. Entries are mean $\pm$ sample standard deviation over three training seeds evaluated on the same 300 test scenarios. Blue and orange headers indicate conventional driving metrics and HDV burden metrics used as satisfaction proxies, respectively. Best means at the reported precision are bold, including ties.}
    \label{tab:main_results}

    \begingroup
    \fontsize{8}{10}\selectfont
    \setlength{\tabcolsep}{3pt}
    \renewcommand{\arraystretch}{1.22}

    \resizebox{\textwidth}{!}{%
    \begin{tabular}{@{}>{\columncolor{methodgreen}}c*{7}{c}@{}}
        \toprule

        \multicolumn{1}{c}{}
        & \multicolumn{4}{>{\columncolor{drivingblue}}c}{
            \textbf{Conventional driving metrics}}
        & \multicolumn{3}{>{\columncolor{burdenorange}}c}{
            \textbf{HDV burden (satisfaction proxies)}} \\

        \cmidrule(lr){2-5}
        \cmidrule(lr){6-8}

        \textbf{Method}
        & \cellcolor{drivingblue!55}
          \mbox{SR $\uparrow$ (\%)}
        & \cellcolor{drivingblue!55}
          \mbox{$\mathrm{TTC}<3\,\mathrm{s}\,\downarrow$ (\%)}
        & \cellcolor{drivingblue!55}
          \mbox{AV time $\downarrow$ ($\mathrm{s}$)}
        & \cellcolor{drivingblue!55}
          \mbox{AV jerk $\downarrow$ ($\mathrm{m/s^3}$)}
        & \cellcolor{burdenorange!55}
          \mbox{Delay $D\downarrow$ ($\mathrm{veh}\cdot\mathrm{s}$)}
        & \cellcolor{burdenorange!55}
          \mbox{Braking $B\downarrow$ ($\mathrm{m/s}$)}
        & \cellcolor{burdenorange!55}
          \mbox{Accel. var. $J\downarrow$ ($\mathrm{m/s^2}$)} \\

        \midrule

        PIWM
        & $\mathbf{100.00}\pm0.00$
        & $63.00\pm17.90$
        & $24.65\pm2.78$
        & $6.896\pm0.877$
        & $19.29\pm4.96$
        & $2.410\pm0.714$
        & $139.95\pm42.42$ \\

        DreamerV3
        & $\mathbf{100.00}\pm0.00$
        & $63.22\pm18.09$
        & $25.20\pm4.05$
        & $2.586\pm3.413$
        & $20.14\pm6.85$
        & $2.461\pm0.750$
        & $89.81\pm59.77$ \\

        CPPO
        & $99.89\pm0.19$
        & $73.67\pm0.00$
        & $23.61\pm0.76$
        & $3.848\pm0.906$
        & $17.92\pm1.90$
        & $2.860\pm0.007$
        & $110.62\pm17.40$ \\

        RD-ACPPO
        & $99.56\pm0.77$
        & $73.67\pm0.00$
        & $23.88\pm1.26$
        & $3.320\pm2.477$
        & $18.08\pm2.22$
        & $2.706\pm0.096$
        & $103.22\pm44.75$ \\

        QR-DQN
        & $\mathbf{100.00}\pm0.00$
        & $73.67\pm0.00$
        & $\mathbf{22.86}\pm0.00$
        & $\mathbf{0.586}\pm0.001$
        & $\mathbf{16.18}\pm0.00$
        & $2.834\pm0.000$
        & $54.93\pm0.00$ \\

        SD-SAC
        & $\mathbf{100.00}\pm0.00$
        & $75.22\pm2.69$
        & $22.98\pm0.19$
        & $0.587\pm0.002$
        & $16.28\pm0.14$
        & $2.765\pm0.118$
        & $54.91\pm0.02$ \\

        \midrule
        \rowcolor[HTML]{F3F7FC}
        \cellcolor{lightmauve!40}\textbf{Ours}
        & $\mathbf{100.00}\pm0.00$
        & $\mathbf{62.11}\pm16.55$
        & $24.24\pm2.39$
        & $0.626\pm0.037$
        & $\mathbf{16.18}\pm0.03$
        & $\mathbf{2.291}\pm0.752$
        & $\mathbf{54.50}\pm0.81$ \\
        \bottomrule
    \end{tabular}%
    }%

    \endgroup
\end{table*}

\subsection{Policy Optimization with Imagination and Execution}
\label{sec:method_policy}

The first policy path evaluates imagined actions using the calibrated critic. From replay-encoded states, the current policy generates $15$-step world-model rollouts and computes
\begin{equation}
\begin{aligned}
\widehat A_t={}&\widehat r_{t+1}
+\gamma\widehat c_{t+1}V_\psi(\widehat x_{t+1})-V_\psi(\widehat x_t),
\end{aligned}
\label{eq:method_imag_adv}
\end{equation}
where $\widehat x_t$, $\widehat r_{t+1}$, and $\widehat c_{t+1}$ are the imagined state, predicted reward, and continuation probability. All rollout positions contribute one-step advantages, while the critic represents subsequent returns and retains multistep supervision. The imagined policy loss is
\begin{equation}
\begin{aligned}
\mathcal L_\pi^{\mathrm{imag}}=-\mathbb E\Bigg[\operatorname{sg}(w_t)\Bigg(&
\log\pi_\theta(\widehat a_t\mid\widehat x_t)
\operatorname{sg}\!\left(\frac{\widehat A_t}{S}\right)\\
&+\frac{\alpha}{S}\mathcal H\!\left[\pi_\theta(\cdot\mid\widehat x_t)\right]\Bigg)\Bigg],
\end{aligned}
\label{eq:method_imag_policy}
\end{equation}
where $w_t$ accumulates discounted continuation, $\operatorname{sg}$ stops gradients, and $\mathcal H$ is policy entropy with coefficient $\alpha$. The detached scale $S$ is the exponentially averaged $95$th--$5$th percentile span of imagined multistep returns, lower-bounded by one.

The second path directly updates executed actions. Reusing $\delta_t$ and $\rho_t$ from Section~\ref{sec:method_value}, we construct
\begin{equation}
A_t^{\mathrm{exec}}
=\delta_t+\gamma(1-d_{t+1})
\lambda\rho_{t+1}A_{t+1}^{\mathrm{exec}}
\label{eq:method_exec_adv}
\end{equation}
over valid segments with known behavior probabilities, with zero subsequent advantage at segment boundaries. The factor $\rho_{t+1}$ weights later actions; the current action weight $\rho_t$ enters the loss:
\begin{equation}
\begin{aligned}
\mathcal L_\pi^{\mathrm{exec}}
={}&-\mathbb E_{\mathrm{valid,known}}\Bigg[
\log\pi_\theta(a_t\mid x_t)\cdot\operatorname{sg}\left(
\frac{\rho_t A_t^{\mathrm{exec}}}{S}
\right)
\Bigg].
\end{aligned}
\label{eq:method_exec_policy}
\end{equation}
The expectation averages supported transitions and shares $S$ with imagined learning. Latent inputs, values, ratios, and advantages are detached, so gradients act through action log-probabilities, reinforcing positive executed advantages and suppressing negative ones.

The two paths are combined as
\begin{equation}
\mathcal L_\pi=
\mathcal L_\pi^{\mathrm{imag}}
+\eta\mathcal L_\pi^{\mathrm{exec}},
\label{eq:method_policy_loss}
\end{equation}
where $\eta=0.1c_{\rm RMS}$ uses a fixed pilot-calibrated scale. Executed outcomes therefore guide both imagined action evaluation and direct policy updates.

\subsection{Joint Training and Execution}
\label{sec:method_training}

Each iteration updates the world model from finalized replay and trains the critic and actor with $\mathcal L_V$ and $\mathcal L_\pi$. Replay selects the latest $20$ completed episodes with probability $0.75$ and the full buffer with probability $0.25$, sampling time steps uniformly within either set. This mixture controls sampling frequency; $\rho_t$ regulates historical actions after sampling.

Training establishes the base model and policy, enables value calibration with one-step imagined advantages, then adds executed policy feedback and the final reward above. A separate pilot from the preceding checkpoint determines $c_{\rm RMS}$ by matching the RMS of weighted executed advantages to a $\lambda=1$ reference on the same data. Final training restores that checkpoint's full model and optimizer state, retaining the calibrated scale and $\lambda=0.95$. Pilot collection counts toward the training budget; its learned weights are discarded. All networks continue to train jointly with stochastic latent states.

At execution, the recurrent state-space model encodes $h_t$ using the posterior mode for its stochastic latent variables. The actor selects $a_t=\arg\max_{a\in\mathcal A}\pi_\theta(a\mid x_t)$ with the learned weights unchanged. Imagination, replay, and feedback calibration remain training-time operations.

\section{Experiment}

\subsection{Experimental Settings}

\textbf{Setup.}
We evaluate our method at an unsignalized intersection in SUMO, with one controlled AV and HDVs approaching from conflicting directions or following the AV. Our policy selects a target speed from $\{0,3,6,9\}\,\mathrm{m/s}$ every $0.1\,\mathrm{s}$ using a 19-step observation history. The passage time limit is $40\,\mathrm{s}$. We use 15-step imagined rollouts, $\beta=0.3$, and $\eta=0.1c_{\mathrm{RMS}}$, where $c_{\mathrm{RMS}}$ is fixed after a separate advantage-scale calibration. Replay samples come from the latest 20 completed episodes with probability $0.75$ and from the full history with probability $0.25$. We evaluate the final policies on the same 300 scenarios and report means and sample standard deviations across three training seeds. Our model is trained on an NVIDIA GeForce RTX 5070 Laptop GPU with 8\,GB of memory.

\textbf{Baselines.}
We compare against six baselines: two world-model methods, PIWM~\cite{gao2024dream} and DreamerV3~\cite{hafner2025mastering}; two PPO-based intersection methods, CPPO~\cite{peng2023curriculum} and RD-ACPPO~\cite{peng2024reward}; and two additional model-free methods, QR-DQN~\cite{dabney2018distributional} and SD-SAC~\cite{wei2025revisiting}. Methods originally developed for different task settings are evaluated using task-adapted implementations.

\textbf{Metrics.}
We report success rate (SR), AV passage time, mean absolute AV jerk, and the fraction of all scenarios with a minimum valid TTC below $3\,\mathrm{s}$. We take the minimum over finite, nonnegative TTC values from SUMO's surrogate safety measure device for conflicts involving the AV. Scenarios without a valid record count as unflagged. Success requires collision-free passage within $40\,\mathrm{s}$, the time assigned to unsuccessful trials. We report HDV costs $D$, $B$, and $J$, with $J_{\mathrm{rear}}$ and $J_{\mathrm{conflict}}$ for following
and conflict-direction HDVs, respectively, to identify where the burden reduction occurs.

\subsection{Performance Comparison and Interaction Analysis}
\label{sec:performance_comparison}

\begin{figure}[!t]
    \centering
    \begingroup
\footnotesize
\setlength{\tabcolsep}{0pt}
\renewcommand{\arraystretch}{0.90}
\begin{tabular}{@{}c@{}c@{}}
(a) Following HDVs & (b) Conflict-direction HDVs \\[1pt]
\includegraphics[page=1,width=.5\linewidth]{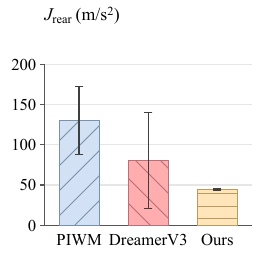} &%
\includegraphics[page=2,width=.5\linewidth]{plot-panels.pdf}
\end{tabular}%
\endgroup%

    \caption{
        HDV acceleration variation for following vehicles (left) and vehicles approaching from conflicting directions (right). Bars and error bars show means and sample standard deviations over three training seeds. The panels use different vertical scales.
    }
    \label{fig:hdv_burden_decomposition}
\end{figure}

Table~\ref{tab:main_results} shows that our method achieves 100\% success and lower mean AV jerk and HDV acceleration variation $J$ than PIWM and DreamerV3. Compared with QR-DQN and SD-SAC, it reduces the fraction of scenarios with $\mathrm{TTC}<3\,\mathrm{s}$ and braking burden $B$, with longer passage times and higher AV jerk. Fig.~\ref{fig:hdv_burden_decomposition} separates $J$ for following and conflict-direction HDVs. Mean $J_{\mathrm{rear}}$ decreases by 65.5\% relative to PIWM and 44.0\% relative to DreamerV3, with less variation across seeds and comparable $J_{\mathrm{conflict}}$. Together with lower AV jerk, the pattern is consistent with smoother AV motion reducing adjustment demands on following HDVs. Relative to DreamerV3, lower passage time, delay $D$, and braking burden $B$ combine smoothness and efficiency. By capturing acceleration changes below the braking threshold, $J$ reveals smoother following responses beyond those measured by $B$.

\subsection{Ablation Study}
\label{sec:ablation}

\begin{table}[!t]
\centering
\caption{Ablation results reported as mean $\pm$ sample standard deviation over three training seeds on the same 300 test scenarios. Best means are bold, including ties. All settings achieve $100\%$ success.}
\label{tab:ablation}
\begingroup
\fontsize{9}{10}\selectfont
\setlength{\tabcolsep}{2pt}
\renewcommand{\arraystretch}{1.15}
\resizebox{\columnwidth}{!}{%
\begin{tabular}{@{}lccccc@{}}
\toprule
\textbf{Method} & $D\downarrow$ $\mathrm{veh}\cdot\mathrm{s}$ & $B\downarrow$ $\mathrm{m/s}$ & $J\downarrow$ $\mathrm{m/s^2}$ & AV time $\downarrow$ $\mathrm{s}$ & AV jerk $\downarrow$ $\mathrm{m/s^3}$ \\
\midrule
\textbf{Full} & $\mathbf{16.18}\pm0.03$ & $\mathbf{2.291}\pm0.75$ & $54.50\pm0.81$ & $24.24\pm2.39$ & $\mathbf{0.626}\pm0.037$ \\
w/o Value & $16.62\pm0.77$ & $2.668\pm0.28$ & $57.549\pm4.54$ & $23.11\pm0.44$ & $0.627\pm0.069$ \\
w/o Policy & $20.18\pm6.92$ & $2.393\pm0.76$ & $54.334\pm6.89$ & $25.23\pm4.08$ & $0.681\pm0.315$ \\
\midrule
$R$ only & $16.55\pm0.63$ & $2.689\pm0.25$ & $58.660\pm6.46$ & $\mathbf{23.08}\pm0.38$ & $0.685\pm0.172$ \\
$M$ only & $20.56\pm6.61$ & $2.402\pm0.77$ & $\mathbf{53.225}\pm5.41$ & $25.40\pm3.94$ & $\mathbf{0.626}\pm0.199$ \\
\bottomrule
\end{tabular}%
}
\endgroup
\end{table}

Table~\ref{tab:ablation} isolates executed value and policy updates through w/o Value and w/o Policy, retaining imagined learning and the full reward. Reward variants retain both calibration paths: $R$ is gated outbound speed recovery; $M$ combines HDV cost reweighting and discounted post-control cost attribution. $R$ only uses the original HDV penalties, while $M$ only omits recovery. Relative to w/o Value, Full lowers $D$, $B$, and $J$ by $2.6\%$, $14.1\%$, and $5.3\%$, with $4.9\%$ longer passage time and nearly unchanged AV jerk. Relative to w/o Policy, Full reduces passage time and $D$ by $3.9\%$ and $19.8\%$, and lowers $B$ and AV jerk by $4.3\%$ and $8.1\%$, while keeping mean $J$ similar ($+0.3\%$). The feedback paths play complementary roles: value calibration favors lower HDV burden, while policy updates improve passage efficiency. Adding $M$ to $R$ reduces $D$, $B$, and $J$ by $2.2\%$, $14.8\%$, and $7.1\%$, with $5.0\%$ longer passage time. Adding $R$ to $M$ reduces passage time and $D$ by $4.6\%$ and $21.3\%$ and lowers $B$, with a $2.4\%$ increase in $J$. HDV cost shaping reduces disturbance, while gated recovery supports progress.

\begin{figure}[!t]
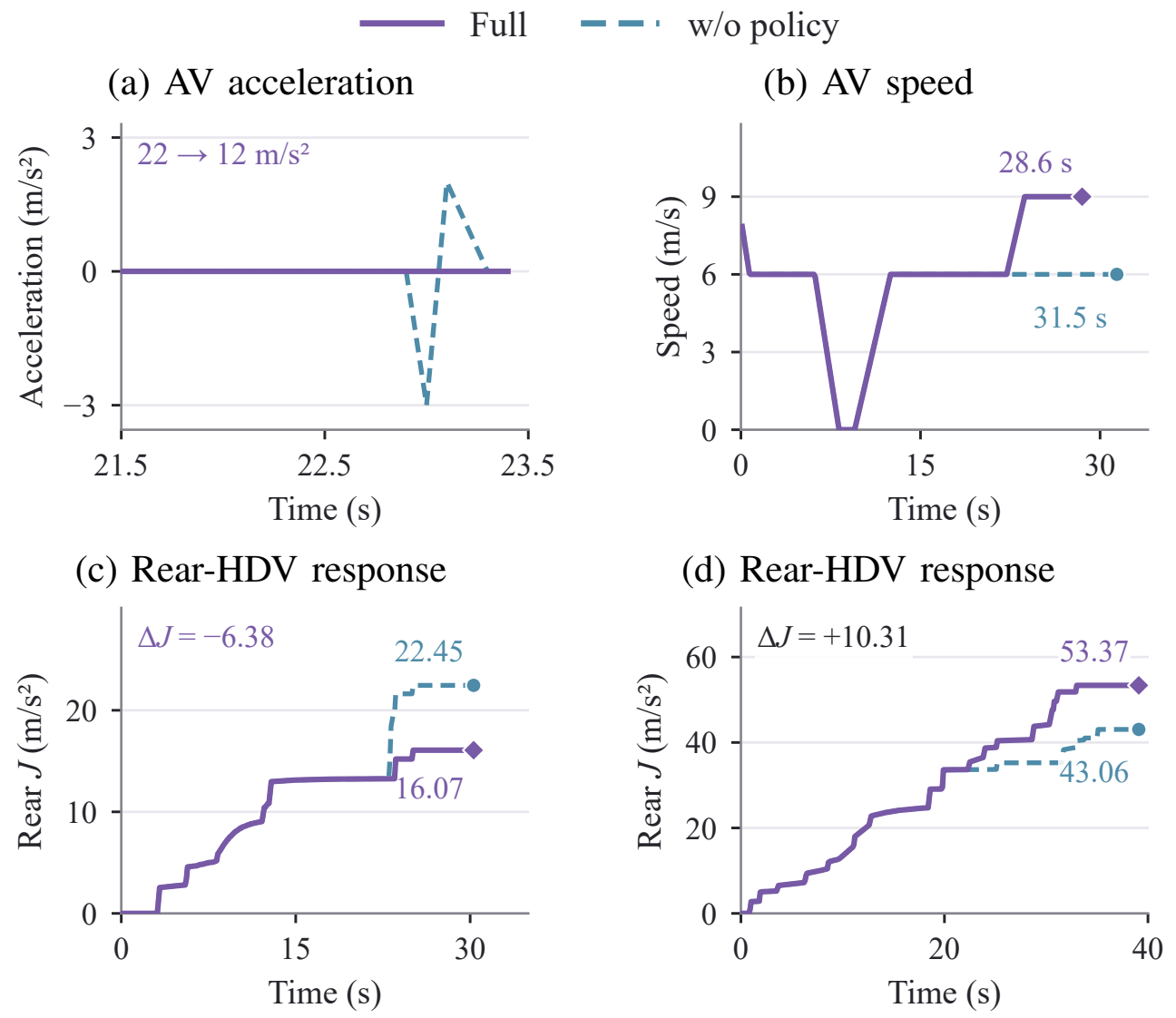

    \centering
    \begingroup
\footnotesize
\setlength{\tabcolsep}{0pt}
\renewcommand{\arraystretch}{0.90}
\begin{tabular}{@{}c@{}c@{}}
\multicolumn{2}{@{}c@{}}{\includegraphics[page=3,width=\linewidth]{plot-panels.pdf}}\\[0.5pt]
(a) AV acceleration & (b) AV speed \\[1pt]
\includegraphics[page=4,width=.5\linewidth]{plot-panels.pdf} &%
\includegraphics[page=5,width=.5\linewidth]{plot-panels.pdf} \\[2pt]
(c) Rear-HDV response & (d) Rear-HDV response \\[1pt]
\includegraphics[page=6,width=.5\linewidth]{plot-panels.pdf} &%
\includegraphics[page=7,width=.5\linewidth]{plot-panels.pdf}
\end{tabular}%
\endgroup%

    \caption{Full versus w/o policy calibration in two illustrative paired cases: (a,c) smoother motion; (b,d) faster passage. Rear-HDV costs cover the full observation window; $\Delta J=J_{\mathrm{Full}}-J_{\mathrm{w/o\ policy}}$.}
    \label{fig:policy_behavior}
\end{figure}

\subsection{Driving Behavior Analysis}
\label{sec:driving_behavior}

The ablation results show shorter average passage time with policy calibration. To identify the driving changes behind this gain, Fig.~\ref{fig:policy_behavior} compares AV motion and following-HDV responses in two illustrative cases, each matched by training seed and test scenario. In panels (a,c), Full avoids a brief braking--acceleration fluctuation, reducing total AV acceleration variation by $45.5\%$ and $J_{\mathrm{rear}}$ by $28.4\%$, while passage time remains $23.5\,\mathrm{s}$. In panels (b,d), Full increases its speed from approximately $6$ to $9\,\mathrm{m/s}$ and exits $2.9\,\mathrm{s}$ earlier ($28.6$ versus $31.5\,\mathrm{s}$). Rear-HDV cost increases from $43.06$ to $53.37\,\mathrm{m/s^2}$, revealing an efficiency--disturbance tradeoff. These cases connect the aggregate results to executed behavior: policy calibration can smooth local motion or shorten passage through a later speed increase, with different consequences for following vehicles.

\begin{figure}[!htbp]
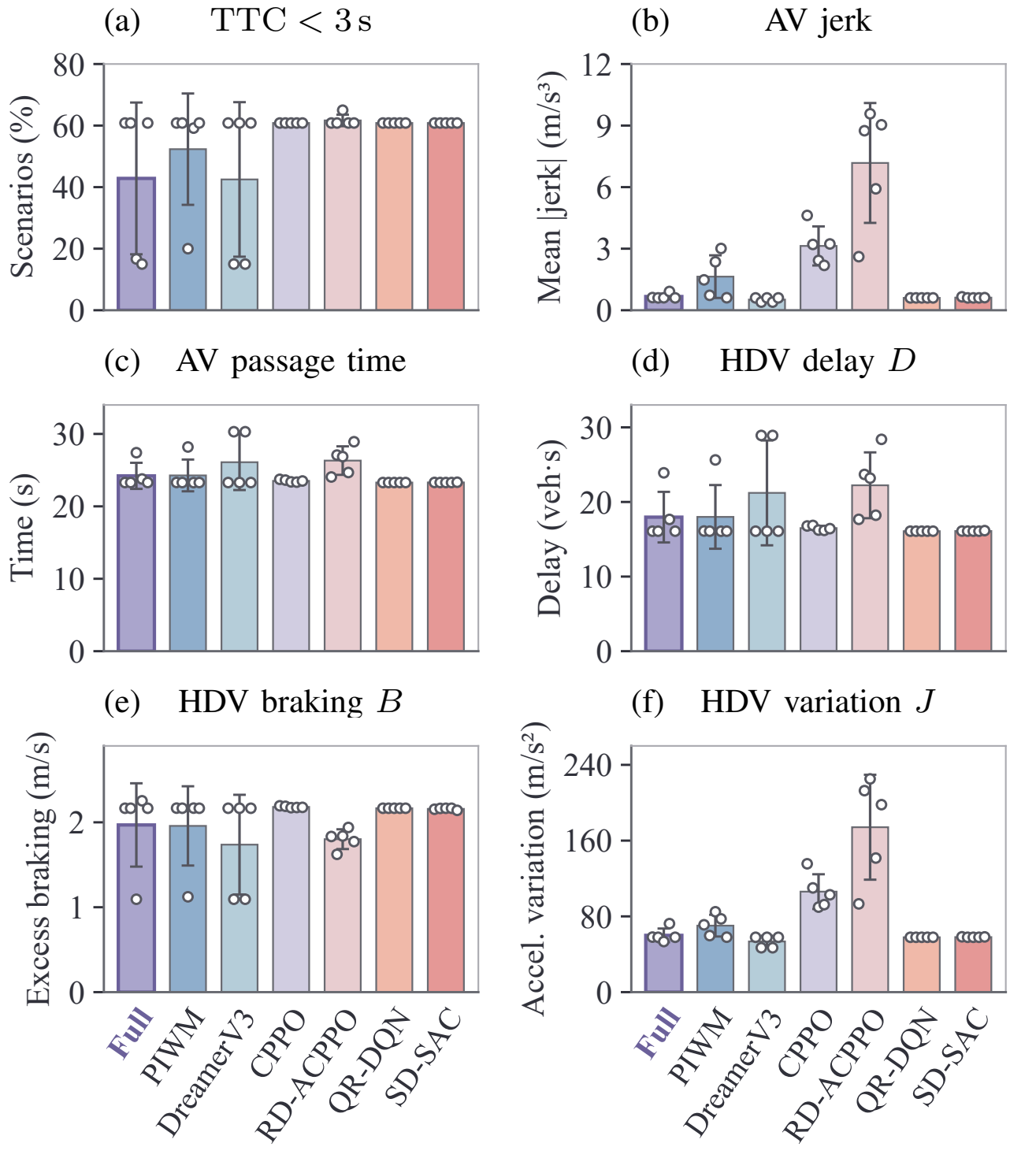

    \centering
    \begingroup
\footnotesize
\setlength{\tabcolsep}{0pt}
\renewcommand{\arraystretch}{0.90}
\newcommand{\oodheading}[2]{%
  \makebox[.5\linewidth][l]{%
    \hspace*{.12\linewidth}\rlap{(#1)}%
    \makebox[.355\linewidth][c]{#2}}}
\begin{tabular}{@{}c@{}c@{}}
\oodheading{a}{$\mathrm{TTC}<3\,\mathrm{s}$} & \oodheading{b}{AV jerk} \\[1pt]
\includegraphics[page=8,width=.5\linewidth]{plot-panels.pdf} &%
\includegraphics[page=9,width=.5\linewidth]{plot-panels.pdf} \\[2pt]
\oodheading{c}{AV passage time} & \oodheading{d}{HDV delay $D$} \\[.66pt]
\includegraphics[page=10,width=.5\linewidth]{plot-panels.pdf} &%
\includegraphics[page=11,width=.5\linewidth]{plot-panels.pdf} \\[2pt]
\oodheading{e}{HDV braking $B$} & \oodheading{f}{HDV variation $J$} \\[-1.1pt]
\includegraphics[page=12,width=.5\linewidth]{plot-panels.pdf} &%
\includegraphics[page=13,width=.5\linewidth]{plot-panels.pdf}
\end{tabular}%
\endgroup%
    \caption{OOD comparison across five training seeds. Bars show means, error bars sample standard deviations, and circles individual seed means.}
    \label{fig:ood_comparison}
\end{figure}

\subsection{Out-of-Distribution Evaluation}
\label{subsec:ood_evaluation}

We evaluate Full and six baselines on the same 120 out-of-distribution (OOD) scenarios using two additional training seeds, for a total of five. Short and long desired HDV headways shift from $[0.90,1.10]$ and $[1.60,1.80]$\,s in training to $[1.15,1.20]$ and $[1.85,1.90]$\,s in testing. The AV starts slower and farther upstream, initial following gaps are set to $20$--$25\,\mathrm{m}$, and later opposing departures are delayed by $1.5$--$3\,\mathrm{s}$. As shown in Fig.~\ref{fig:ood_comparison}, Full achieves mean passage time and HDV delay comparable to PIWM while reducing ego jerk and HDV acceleration variation $J$ by 59.2\% and 14.6\%, respectively, with consistent reductions across all five corresponding seeds. This demonstrates reduced motion disturbance for both the ego vehicle and surrounding HDVs at similar passage efficiency. Compared with QR-DQN and SD-SAC, Full yields an 18-percentage-point lower mean fraction of scenarios with TTC below 3\,s and approximately 9\% lower braking cost $B$, accompanied by about 0.9\,s longer mean passage time. These results reveal a trade-off between lower conflict exposure and braking burden and faster passage in the evaluated OOD conditions.

\section{Conclusion}
\label{sec:conclusion}

This paper developed a world-model reinforcement learning framework that calibrates value and policy learning with finalized interaction feedback. By retaining HDV costs that arise after AV control ends, this feedback connects both learning paths to the consequences of executed interactions. At an unsignalized intersection, the method achieves 100\% success and lower AV jerk and HDV acceleration variation than PIWM and DreamerV3, primarily benefiting following traffic. Ablations link value calibration to lower HDV burden and policy calibration to faster passage, supporting executed feedback as a complement to imagined learning. OOD results further show smoother ego and HDV motion than PIWM at comparable passage times. Calibration remains a training-time operation, preserving direct policy execution without online imagined rollouts.

\bibliographystyle{IEEEtran}
\bibliography{ref}
\end{document}